\documentclass{article}

\usepackage{arxiv}

\usepackage[utf8]{inputenc}
\usepackage[T1]{fontenc}
\usepackage{url}
\usepackage{booktabs}
\usepackage{amsfonts}
\usepackage{nicefrac}
\usepackage{microtype}
\usepackage{graphicx}
\usepackage{natbib}
\usepackage{doi}
\usepackage{multirow}
\usepackage{multicol}
\usepackage{rotating}

\title{The promising potential of vision language models for the generation of textual weather forecasts}
\date{}

\author{Edward C. C. Steele\thanks{{\texttt{edward.steele@metoffice.gov.uk}}} \\
	Met Office \\
	\And
	Dinesh Mane \\
	Amazon Web Services\\
    \And
	Emilio Monti \\
	Amazon Web Services\\
    \And
	Luis Orus \\
	Amazon Web Services\\
    \And
	Rebecca Chantrill-Cheyette\\
	University of East Anglia\thanks{{Work completed while at the Met Office}}\\
    \And
	Matthew Couch\\
	Amazon Web Services\\
    \And
	Kirstine I. Dale\\
	Met Office\\
    \And
	Simon Eaton\\
	Met Office\\
    \And
	Govindarajan Rangarajan\\
	Amazon Web Services\\
    \And
	Amir Majlesi\\
	Amazon Web Services\\
    \And
	Steven Ramsdale\\
	Met Office\\
    \And
	Michael Sharpe\\
	Met Office\\
    \And
	Craig Smith\\
	Amazon Web Services\\
    \And
	Jonathan Smith\\
	Met Office\\
    \And
	Rebecca Yates\\
	Met Office\\
    \And
	Holly Ellis\\
	Amazon Web Services\\
    \And
	Charles Ewen\\
	Met Office\\
}

\renewcommand{\shorttitle}{\textit{The promising potential of vision language models}}

\begin{document}
\maketitle

\begin{abstract}
	Despite the promising capability of multimodal foundation models, their application to the generation of meteorological products and services remains nascent. To accelerate aspiration and adoption, we explore the novel use of a vision language model for writing the iconic Shipping Forecast text directly from video-encoded gridded weather data. These early results demonstrate promising scalable technological opportunities for enhancing production efficiency and service innovation within the weather enterprise and beyond.
\end{abstract}

\section*{Main}

We are presently experiencing a revolution in artificial intelligence (AI), with pioneering developments in machine learning weather prediction (MLWP) models demonstrating skill comparable to physics-based numerical weather prediction (NWP) models across a range of attributes at substantially lower computational cost \citep{lam_etal_2023, allen_etal_2025, bodnar_etal_2025}. Despite these advances, most of the early AI applications within the meteorological discipline have focused on tasks related to the generation of raw weather \textit{predictions} rather than on the generation of weather \textit{products}. This is significant as a weather forecast, however skillful, has no intrinsic value unless its user can derive a benefit from it \citep{mylne_2002}. Consequently, it is reasonable to expect the benefit of alleviating the similarly resource-intensive effort typically required for the production of purposeful textual forecast bulletins, narratives and warnings that are useful, usable and used will likely match -- if not exceed -- the impact of AI use in other parts of the science-to-services value chain. While simple rules-based approaches provide an alternative means of automation in instances where data-to-text conversion is able to be coded explicitly, to enable wider product and service transformation on an enterprise scale it is essential to develop and improve solutions in a more efficient and effective manner that doesn’t cause an unnecessary proliferation of isolated microservices. Against the background of an ever-expanding demand from forecast consumers for the multimodal provision of personalized weather data/intelligence across an increasing range of required applications, the advent of Vision Language Models (VLMs; \citet{bordes_etal_2024}) -- that extend the capabilities of Large Language Models (LLMs; \citet{vaswani_etal_2023, minaee_etal_2025}) by combining computer vision and natural language processing -- offer new possibilities for the scalable generation of textual meteorological products and services directly from gridded NWP or MLWP model output. Indeed, the increased availability of foundation models, combined with the opportunities for leveraging the latest advances more accessibly, afford new approaches for extracting meaning from complex scientific data -- with LLMs/VLMs offering a different way to deliver information in a less labor-intensive manner -- potentially leading to a whole new generation of products and services. In this paper we are, therefore, motivated to share our experiences from a targeted prototyping activity exploring the novel use of a VLM fine-tuned for meteorological data-to-text conversion using the writing of the iconic Shipping Forecast as an example.

Celebrating its BBC broadcast centenary in 2025, the Shipping Forecast (the oldest and longest running weather forecast in the world) is a British cultural institution and cornerstone of regional maritime safety. Issued by the Met Office on behalf of the Maritime \& Coastguard Agency (MCA), the forecast requires that the predicted conditions for each of its 31 constituent sea areas out to 24 hours ahead are analyzed and condensed into text sentences with a very specific length and format according to a strict set of rules. The output contains a general (pressure) synopsis followed by the area bulletins themselves. These use an 8-point compass for the wind direction, the Beaufort Scale for the wind strength, the Douglas Scale for wave height, a four-category classification of visibility and a maximum of five words to highlight any high impact weather, with key timing phrases within the final text breaking the 24-hour period into sub-periods. Finally, sea areas are grouped together to avoid repetition.

Drawing on a combination of existing atmospheric and oceanographic NWP data, as well as a complementary archive of manually generated and issued text bulletins, the Shipping Forecast provides an ideal test case for meteorological data-to-text conversion, representative of a broader set of related/similar tasks within the weather enterprise and beyond.

To overcome the token-size constraints associated with attempting to convey temporal forecast information to a VLM alongside individual images, the gridded NWP data were encoded in video format for direct vision processing using Amazon Nova Lite 1.0 models \citep{amazon_agi_2025} -- with this work representing the first example of their custom fine-tuning for optimizing performance in application (Table~\ref{tab:table1}).

\begin{sidewaystable}[p]
\caption{Experiments optimizing VLM performance. Note that the evaluations presented employ a particularly strict word-level scoring which requires an exact match to the text generated by Met Office meteorologists.}
\centering
\begin{tabular}{ccclccc}
\toprule
\multicolumn{2}{l}{\textbf{Attribute(s)}}                                              & \textbf{Experiment}   & \textbf{Configuration}                  & \textbf{Precision} & \textbf{Recall} & \textbf{Word F1 Score} \\
\midrule
\multirow{3}{*}{Combined}    & \multirow{3}{*}{Sea State, Visibility, Weather \& Wind} & Exp1                  & Continuous                               & 52.8\%          & 28.6\%       & 35.6\%                    \\
                             \cmidrule{3-7}
                             &                                                         & Exp2                  & Categorical (30 epochs)                  & 52.2\%          & 32.4\%       & 38.3\%                    \\
                             \cmidrule{3-7}
                             &                                                         & Exp3                  & Categorical (early stop)                 & 51.8\%          & 29.3\%       & 36.1\%                    \\
                             \cmidrule{1-7}
\multirow{17}{*}{Individual} & \multirow{3}{*}{Pressure*}                              & Exp1                  & N/A                                      & N/A          & N/A       &  N/A                   \\
                             \cmidrule{3-7}
                             &                                                         & Exp2                  & Categorical (30 epochs)                  & 25.2\%          & 17.5\%       & 20.0\%                    \\
                             \cmidrule{3-7}
                             &                                                         & Exp3                  & Categorical (early stop – 10 epochs)     & 27.9\%          & 21.8\%       & 23.6\%                    \\
                             \cmidrule{2-7}
                             & \multirow{3}{*}{Sea State}                              & Exp1                  & Continuous                               & 77.6\%          & 38.2\%       & 46.5\%                    \\
                             \cmidrule{3-7}
                             &                                                         & Exp2                  & Categorical (30 epochs)                  & 88.7\%          & 43.4\%       & 53.4\%                    \\
                             \cmidrule{3-7}
                             &                                                         & Exp3                  & Categorical (early stop – 21 epochs)     & 91.4\%          & 39.4\%       & 50.4\%                    \\
                             \cmidrule{2-7}
                             & \multirow{3}{*}{Visibility}                             & Exp1                  & Continuous                               & 87.7\%          & 41.0\%       & 51.2\%                    \\
                             \cmidrule{3-7}
                             &                                                         & Exp2                  & Categorical (30 epochs)                  & 89.3\%          & 49.3\%       & 58.4\%                    \\
                             \cmidrule{3-7}
                             &                                                         & Exp3                  & Categorical (early stop – 21 epochs)     & 90.9\%          & 45.4\%       & 55.4\%                    \\
                             \cmidrule{2-7}
                             & \multirow{3}{*}{Weather}                                & Exp1                  & Continuous                               & 25.7\%          & 19.8\%       & 21.2\%                    \\
                             \cmidrule{3-7}
                             &                                                         & Exp2                  & Categorical (30 epochs)                  & 51.2\%          & 37.5\%       & 40.5\%                    \\
                             \cmidrule{3-7}
                             &                                                         & Exp3                  & Categorical (early stop – 18 epochs)     & 45.0\%          & 29.7\%       & 33.4\%                    \\
                             \cmidrule{2-7}
                             & \multirow{5}{*}{Wind}                                   & Exp1                  & Continuous                               & 60.8\%          & 28.1\%       & 36.6\%                    \\
                             \cmidrule{3-7}
                             &                                                         & Exp2                  & Categorical (30 epochs)                  & 59.4\%          & 36.5\%       & 42.7\%                    \\
                             \cmidrule{3-7}
                             &                                                         & Exp3                  & Categorical (early stop – 18 epochs)     & 54.1\%          & 31.5\%       & 37.6\%                    \\
                             \cmidrule{3-7}
                             &                                                         & \multirow{2}{*}{Exp4} & Full Rank / SFT (categorical, 20 epochs) & 55.9\%          & 47.5\%       & 48.6\%                    \\
                             \cmidrule{4-7}
                             &                                                         &                       & LoRA / PEFT (categorical, 20 epochs)     & 59.4\%          & 36.5\%       & 42.7\%                    \\
                             \bottomrule
\end{tabular}
\label{tab:table1}
\end{sidewaystable}

Despite the initial attractiveness of adopting a combined approach with the aim of maximizing single-inference efficiency, the application of multiple, specific, individual models was found to outperform a more generic fine-tuning strategy. This is largely due to greater control -- allowing the potential for independent optimization of each attribute through specific training epochs, mappings, metrics and prompts -- while their modular architecture permits a greater flexibility for updates and implementation.

Subsequent improvements were achieved by converting the image color-scale from continuous values (Exp1) to categorical labels (Exp2) to better match those used within the Shipping Forecast; with the largest gain seen for the weather type, owing to challenges in the VLM interpreting numeric colorbar information in the context of its prompt. Of the individual attributes considered, categorized visibility performed best (F1: 58.4\%), having the clearest (and fewest) categorical boundaries, while pressure performed worst (F1: 20.0\%) due to the large-scale complexity being less well-constrained. Somewhat counterintuitively, parallel experiments revealed that -- with the exception of pressure -- training the categorical models for 30 epochs (Exp2) consistently outperformed simulations of early stopping (Exp3) usually employed to prevent overfitting, despite these having a higher validation loss. This is attributed to the unique demands of the Shipping Forecast (e.g. limited vocabulary, specific format and exact word matching), as overfitting enhances memorization of precise word patterns and produces more confident outputs (improving word-level F1 scores by generating more accurate and complete forecasts), since the fine-tuning experiments were constrained to optimizing based on token-level probabilities, or ‘perplexity’, rather than the target metric itself. In specialized domains such as weather forecasting, monitoring both training and evaluation metrics to identify optimal stopping points is therefore essential, particularly if no opportunity for providing a custom loss function exists.

To supplement these learnings, with a view to possible future enhancement, an additional comparison between full rank Supervised Fine-Tuning (SFT) and Parameter-Efficient Fine-Tuning (PEFT) using Low Rank Adaptation (LoRA; \citet{hu_etal_2021}) for the wind attribute model was conducted. Exp4 revealed that under identical conditions (20 epochs, categorical) full rank fine-tuning achieved a 5.9\% relative improvement in word-level F1 score, with improved recall indicating more accurate and complete forecast generation. This performance boost is attributed to full rank updates being applied to all model parameters, enabling better learning of complex meteorological patterns and formatting precision. While LoRA offers greater efficiency and lower computational cost, the superior accuracy and control of full rank fine-tuning could make it the preferred choice for production, with the trade-off in longer training time ($\times$12-15) and higher deployment cost (using Provisional Throughput rather than On-Demand capacity) justified by the gains in forecast quality. However, this is likely only true for highly specialized applications, such as the Shipping Forecast, since LoRA not only learns less, but also forgets less -- i.e. if the model has to perform only a single narrow task, full rank fine-tuning can still have a performance advantage, but with full rank fine-tuning the model can catastrophically forget other capabilities not represented in the fine-tuning training set \citep{biderman_etal_2024}. That said, emergent research \citep{schulman_2025} suggests that applying LoRA to all layers, rather than just the attention matrices, and increasing the learning rate by a factor of 10, might instead offer a regret-free approach to fine-tuning that should be considered in the future.

Compared to an LLM baseline (based on Amazon Nova Pro 1.0) that processed data through an intermediate text representation stage prior to generating forecasts, in a separate experiment, the use of the VLM revealed distinct strengths and limitations (Table~\ref{tab:table2}). The VLM showed higher precision for sea state (+11\%), visibility (+19\%) and weather (+11\%), indicating more accurate word generation. However, this came at the cost of much lower recall, suggesting the video-based approach was more conservative in its predictions. Notably, the VLM slightly outperformed the LLM in weather type identification, benefiting from visual pattern recognition, however the wind forecast generation was found to be more challenging for the VLM possibly due to the complexity of integrating both direction and speed from short video segments.

\begin{table}[h]
	\caption{Evaluation of VLM performance against LLM baseline.}
	\centering
    \begin{tabular}{clccc}
    \toprule
    \multicolumn{1}{c}{\textbf{Attribute(s)}} & \textbf{Word Metric} & \textbf{LLM} & \textbf{VLM} & \textbf{Difference} \\
    \midrule
    \multirow{3}{*}{Sea State}        & Precision     & 78\%    & 89\%    & -11\%           \\
                                      \cmidrule{2-5}
                                      & Recall        & 73\%    & 44\%    & 29\%           \\
                                      \cmidrule{2-5}
                                      & F1            & 75\%    & 59\%    & 16\%           \\
                                      \cmidrule{1-5}
    \multirow{3}{*}{Visibility}       & Precision     & 70\%    & 89\%    & -19\%           \\
                                      \cmidrule{2-5}
                                      & Recall        & 80\%    & 50\%    & 30\%           \\
                                      \cmidrule{2-5}
                                      & F1            & 75\%    & 64\%    & 11\%           \\
                                      \cmidrule{1-5}
    \multirow{3}{*}{Weather}          & Precision     & 40\%    & 51\%    & -11\%           \\
                                      \cmidrule{2-5}
                                      & Recall        & 44\%    & 38\%    & 6\%           \\
                                      \cmidrule{2-5}
                                      & F1            & 42\%    & 44\%    & -2\%           \\
                                      \cmidrule{1-5}
    \multirow{3}{*}{Wind}             & Precision     & 58\%    & 53\%    & 5\%           \\
                                      \cmidrule{2-5}
                                      & Recall        & 57\%    & 32\%    & 25\%           \\
                                      \cmidrule{2-5}
                                      & F1            & 57\%    & 40\%    & 17\%           \\
                                      \cmidrule{1-5}
    \textbf{Average}                  & \textbf{F1}            & \textbf{62\%}    & \textbf{52\%}    & \textbf{10\%}    \\
    \bottomrule
	\end{tabular}
	\label{tab:table2}
\end{table}

Overall, across the four core weather attributes included within the sea area bulletins, the LLM achieved an average word-level F1 score of 62\% -- outperforming the score of 52\% achieved by the VLM -- in part, since we were able to apply extra domain knowledge to enhance its input. However, as textual data descriptions represent an information bottleneck, with continued development we expect the VLM to ultimately be capable of outperforming the LLM on this task. Nonetheless, while still at an early stage of exploration, these results demonstrate the real potential for simplifying the task of meteorological data-to-text conversion -– as well as approaches for extracting meaning from complex scientific data more generally. Suggested opportunities for increasing performance include expansion of training datasets, enhancement of prompt engineering (e.g. including more structured spatial/temporal guidance, example patterns and definitions of domain-specific terminology), and exploration of larger models such as Amazon Nova Pro 1.0; with hybrid applications (combining text and video input) also worth being considered. This adoption of pioneering generative AI methods is undoubtedly a key future enabler of products and services transformation and, if a similar rate of progress is made for these as seen for MLWP, then we can anticipate the realization of these benefits in the very near future.

\section*{Methods}

The VLM fine-tuning architecture was implemented using Amazon SageMaker to support distributed training across high-performance cloud-based compute instances. The complete pipeline enabled a workflow comprising data preparation/processing, model training, and solution deployment. For data-to-text conversion, gridded atmosphere and ocean predictions, representing the required elements of the Shipping Forecast, were first converted into short video segments, prior to pairing with its corresponding (archived) text bulletin, to enable visual learning.

The meteorological data consisted of single-level deterministic fields of weather type and wind direction, and probabilistic fields of pressure, visibility and wind speed from the global IMPROVER system \citep{roberts_etal_2023}, while the sea state data consisted of single-level deterministic significant wave height fields from the global ocean wave model \citep{valiente_etal_2023}, each with an approximate spatial resolution of 20 km and cropped to a regional domain described by the coordinates: \{30°N, 70°N\}, \{-20°E, 10°E\}. For simplicity, all probabilistic fields presented to the VLM were based on the 50th percentile of the ensemble distribution \citep{mylne_etal_2025}. For each  individual sea area prediction used, each weather attribute was encoded as a 1-second video (2-second video for combined wind direction and wind speed attributes) at  24 frames per second, with each frame representing one-hourly forecast timestep, covering all 24 hours of the day. These were stored in MOV format and standardized to a $10\times6$ figure aspect ratio to ensure consistent dimensions. Due to its synoptic nature, pressure was processed separately using a distinct strategy that involved overlaying a labeled grid, rather than masking individual sea areas (as employed otherwise). The combined corpus comprises approximately 1,500 video-bulletin pairings across a three-month period from December 2024 to February 2025 (inclusive), used for more rapid experimental iteration, and employed a random shuffle with a 70\%:15\%:15\% training/validation/test data split. Unless specified, the fine-tuning employed Low-Rank Adaptation (LoRA) of the Amazon Nova Lite 1.0 model (a state-of-the-art foundation model capable of video input), enabling efficient adjustment of model weights while minimizing computational overhead. Training was distributed across four ml.p5.48xlarge instances, each equipped with $8\times$NVIDIA H100 GPUs. A complete training run spans 30 epochs and typically completed within 60-90 minutes. To accelerate development, up to five parallel fine-tuning jobs were executed simultaneously, each targeting a different weather attribute. Upon completion, adapter weights were stored in Amazon S3 and deployed to custom endpoints on Amazon Bedrock for inference via the Amazon Bedrock Converse API, enabling the scalable real-time generation of the shipping forecast text from gridded meteorological data.

The LLM baseline comparisons leveraged a pre-trained foundation model, where the Shipping Forecast generation was framed as a set of LLM tasks (e.g. attribute generation, gale warning generation, general synopsis generation and sea area consolidation); each defined by a system prompt (forecast guidance), user prompt (rules-based textual representation of the input gridded weather data) and 21 few-shot examples, as well as model selection and temperature settings. In contrast to the VLM, all probabilistic fields presented to the LLM consisted of basic macro-statistics extracted from the 5th and 85th percentile of the ensemble distribution, as determined from a previous Met Office calibration, in an attempt to apply extra domain knowledge to enhance its input (albeit recognizing the capacity for summarizing the characteristics of the input datasets in this way is limited). Although the use of both Amazon Nova Pro 1.0 and Anthropic Claude 3.7 Sonnet models were tested for the LLM baseline, the former was chosen as it achieved a 5\% higher word-level F1 score across the four core weather attributes included within the sea area bulletins, with a low temperature (0.2) used to increase precision.

Where possible, both the LLM and VLM prompts imported the detailed guidance written for the human meteorologists, to assess the extent that pre-existing Met Office work instructions and documentation can be deployed directly. Conversions from continuous to categorical values used by the LLM, and in the categorical experiments by the VLM, similarly classified data into bins aligned with the specific vocabulary of the Shipping Forecast \citep{metoffice_2025}.

To evaluate performance, and consistent with the training/inference approaches, the Shipping Forecast was segmented by sea-area and further split into attributes. Forecasts covering multiple sea-areas were excluded, as the model generates text per sea-area, with aggregation handled separately. Due to its highly specific and structured nature, LLM-based evaluation was deemed unsuitable. Instead, a very strict word-level comparison that counts the matching words (true positives), missing words (false negatives), and extra words (false positives) between the generated text and the expected text was used, with aggregate scores across tests computed as the micro-averaged F1 score by averaging precision and recall. As an alternative to word-level precision and recall we also considered the BERT score \citep{zhang_etal_2020}, however this provides a softer match between words -- resulting in an overly lenient impression of performance in applications where extreme precision is required -- and so was discounted.

\section*{Data availability}

The Shipping Forecast bulletins were obtained from the UK National Meteorological Library and Archive (Exeter, UK). The accompanying post-processed atmosphere and ocean model used can be made available, on request, under the terms of a Met Office data license for non-commercial research use.

\bibliographystyle{abbrvnat}
\setcitestyle{authoryear,open={((},close={))}}
\bibliography{steele_etal_2025_references}

\section*{Acknowledgments}

The Shipping Forecast is produced by the Met Office on behalf of the Maritime \& Coastguard Agency, whose permission to use these archived bulletins for the prototype -- as well as their ongoing encouragement for harnessing technological developments to drive continuous improvements -- is much appreciated. The authors are grateful to Met Office colleagues Avalon Anglin-Jaffe, Simon Brown, Steven Calder, Matthew Dagnall, Daryl Edwards, Richard Lawrence, Graham Mallin, William Maunder, Rachel McInnes, Edward Pope, Joseph Sach, Andrew Saulter and Jessica Standen, as well as Amazon Web Services colleagues Roland Barcia, Anupam Dewan, Toshal Dudhwala and Kandha Sankarapandian, for their ad hoc input and advice during the project.

\end{document}